\documentclass[letterpaper]{article} 
\usepackage{aaai25}  
\usepackage{times}  
\usepackage{helvet}  
\usepackage{courier}  
\usepackage[hyphens]{url}  
\usepackage{graphicx} 
\usepackage{natbib}  
\usepackage{caption} 
\usepackage{algorithm}
\usepackage{algorithmic}
\usepackage{amsmath}

\usepackage{newfloat}
\usepackage{listings}
\DeclareCaptionStyle{ruled}{labelfont=normalfont,labelsep=colon,strut=off} 
\floatstyle{ruled}
\newfloat{listing}{tb}{lst}{}
\floatname{listing}{Listing}
\usepackage{hyperref}   
\usepackage{subcaption}
\usepackage{xcolor}
\usepackage{booktabs}
\usepackage{cleveref}
\usepackage{xspace}

\newcommand{\ours}{ALaST\xspace}
\newcommand{\deits}{DeiT-S\xspace}
\newcommand{\deitt}{DeiT-T\xspace}
\newcommand{\vit}{ViT-B\xspace}

\newcommand{\flowers}{Flower-102\xspace}
\newcommand{\food}{Food-101\xspace}
\newcommand{\cifar}{Cifar-100\xspace}

\newcommand{\citenump}[1]{[\citenum{#1}]}

\definecolor{darkblue}{rgb}{0, 0, 0.5}
\hypersetup{colorlinks=true, citecolor=darkblue, linkcolor=darkblue, urlcolor=darkblue}

\title{Adaptive Layer Selection for \\ Efficient Vision Transformer Fine-Tuning}

\author{
    Alessio Devoto\textsuperscript{\rm 1 \rm 3}
    Federico Alvetreti\textsuperscript{\rm 1}
    Jary Pomponi\textsuperscript{\rm 1 \rm 3}\\
    Paolo Di Lorenzo\textsuperscript{\rm 1 \rm 3}
    Pasquale Minervini\textsuperscript{\rm 2}
    Simone Scardapane\textsuperscript{\rm 1 \rm 3}
    }
\affiliations {
    \textsuperscript{\rm 1} La Sapienza, University of Rome \\
    \textsuperscript{\rm 2} The University of Edinburgh 
    \textsuperscript{\rm 3} CNIT \\
    \textsuperscript{\rm 1} \{firstname.lastname\}@uniroma1.it 
    \textsuperscript{\rm 2} p.minervini@ed.ac.uk
}

\begin{document}

\maketitle

\begin{abstract}
Recently, foundation models based on Vision Transformers (ViTs) have become widely available. However, their fine-tuning process is highly resource-intensive, and it hinders their adoption in several edge or low-energy applications. To this end, in this paper we introduce an efficient fine-tuning method for ViTs called \textbf{ALaST} (\textit{Adaptive Layer Selection Fine-Tuning for Vision Transformers}) to speed up the fine-tuning process while reducing computational cost, memory load, and training time. Our approach is based on the observation that not all layers are equally critical during fine-tuning, and their importance varies depending on the current mini-batch. Therefore, at each fine-tuning step, we adaptively estimate the importance of all layers and we assign what we call ``compute budgets''  accordingly. Layers that were allocated lower budgets are either trained with a reduced number of input tokens or kept frozen.
Freezing a layer reduces the computational cost and memory usage by preventing updates to its weights, while discarding tokens removes redundant data, speeding up processing and reducing memory requirements.
We show that this adaptive compute allocation enables a nearly-optimal schedule for distributing computational resources across layers, resulting in substantial reductions in training time (up to 1.5x), FLOPs (up to 2x), and memory load (up to 2x) compared to traditional full fine-tuning approaches. Additionally, it can be successfully combined with other parameter-efficient fine-tuning methods, such as LoRA. 
\end{abstract}

%

\section{Introduction}

Recently, large-scale Vision Transformers (ViTs, \cite{dosovitskiy2021vit}) have become the leading paradigm in computer vision. By leveraging the self-attention mechanism, ViTs can capture long-range dependencies in images at every layer, leading to superior results compared to traditional convolutional neural networks (CNNs). ViTs are at the core of a wide array of applications, ranging from vision-language models \cite{clip,owlvit,llava} to resource-constrained embedded devices \cite{cai2022efficientvit, tinytl, vit_embedded_apple, vit_device}.

\begin{figure}[t]
    \centering
    \includegraphics[width=0.9\linewidth]{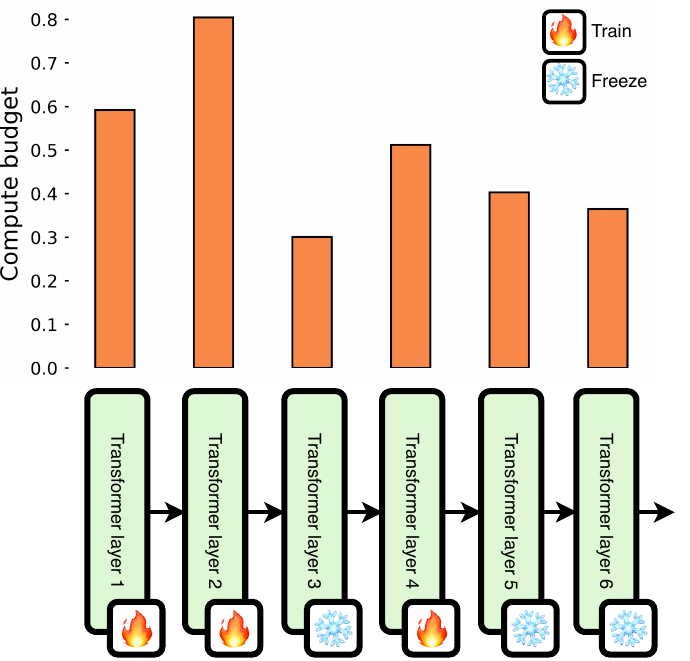}
    \caption{At each fine-tuning step, we assign what we call ``compute budgets'' to transformer layers. The budget determines the computational resources we invest in each layer, i.e., (a) whether the layer is frozen or trainable and (b) how many tokens that layer can process. By adaptively allocating the budget, we make the fine-tuning faster and more efficient in terms of FLOPs, memory, and time.}
    \label{fig:first_page}
\end{figure}

The impressive capabilities of ViTs come with the drawback of a resource-intensive training process. The high computational demands arise from the large number of parameters and the quadratic complexity of the self-attention mechanism. Moreover, ViTs are extremely data-hungry, requiring large datasets to achieve optimal performance, which in turn prolongs training times. 

To address these constraints, a common practice for the deployment of ViTs is to leverage a pre-trained foundation model and then perform fine-tuning for specific tasks. By updating only a subset of the model parameters, fine-tuning makes it feasible to achieve high performance on specialized tasks without the prohibitive costs associated with training a model from scratch \cite{xin2024parameterefficientfinetuningpretrainedvision}. However, fine-tuning ViTs introduces additional complexity: the choice of which parameters to update is critical, as it can significantly impact performance \cite{bsr}. Identifying the optimal layers to fine-tune often requires extensive experimentation, which is not always feasible in real-world scenarios due to time and computational constraints.
Parameter-efficient fine-tuning (PEFT) methods, such LoRA \cite{hu2022lora} or adapters \cite{adapters} aim to address some of these challenges, but they often focus on optimizing the count of additional parameters rather than reducing overall computational load. As a result, PEFT methods might still be unsuitable for highly constrained environments where minimizing computation and memory are limited \cite{bsr}. 
This is the case for mobile and edge devices, drones or next-generation 6G networks that face strict constraints on memory and computational resources \cite{cai2022efficientvit, tinytl, 6g}. These devices require lightweight models that can be quickly fine-tuned on a low budget without compromising performance. Similarly, in situations where privacy concerns prevent data from being sent to a server, on-device processing and fine-tuning become essential.

In this work, we propose a simple method to accelerate the fine-tuning of ViTs in low-resource settings by leveraging two key insights. First, it is known that not all \emph{tokens} are equally useful during training, due to redundant information present in input images. Various techniques have been developed to estimate token importance and either discard, merge, or halt redundant tokens to enhance \textit{inference} (not training) speed, showing promising results \cite{meng2022adavit, bolya2022tokenmerging, rao2021dynamicvit}. Second, not all \emph{layers} are equally important during training. Some layers receive minimal updates due to small gradients, making associated computation inefficient in both the forward and backward pass.
Building on these observations, we propose \emph{Adaptive LAyer Selective fine-Tuning for Vision Transformers} (\ours) --- during fine-tuning, we allocate a so-called scalar \textit{budget} to each layer, and we control resource consumption along two axes: the number of discarded tokens and the selection of trainable layers. Specifically, we \emph{adaptively} determine (a) how many tokens to forward through each layer and (b) which layers to freeze based on the budget allocation.
Discarding tokens significantly reduces FLOPs -- due to the quadratic cost of multi-head attention -- and accelerates training, especially on mid-range GPUs, enabling models to reach higher accuracy in shorter time. Freezing layers enhances energy efficiency and substantially reduces memory usage, which is critical for on-device training. 

We are the first, to the best of our knowledge, to introduce an adaptive framework that systematically optimizes both token and layer selection during fine-tuning, addressing the computational challenges of ViTs in resource-constrained environments.  We validate our method against a comprehensive set of baseline approaches, ensuring that our results are robust and demonstrating that \ours achieves superior efficiency while maintaining competitive accuracy.
\section{Background on Vision Transformer}
A Vision Transformer (ViT) processes an image \( X \in \mathcal{R}^{C \times H \times W} \) (where $C, H, W$ are channels, width, and height respectively) through a series of \( L \) transformer layers. The transformation pipeline can be formalized as follows:
\begin{equation}
    y = \mathcal{C} \circ \mathcal{F}^{L} \circ \mathcal{F}^{L-1} \circ \dots \circ \mathcal{F}^1 \circ \mathcal{E}(X),
\end{equation}
where \( \mathcal{E}(\cdot) \) denotes the encoding network, \( \mathcal{F}^i \) represents the \( i \)-th transformer layer, and \( \mathcal{C}(\cdot) \) is the classification head.
The encoding network \( \mathcal{E}(\cdot) \) splits the image into smaller, non-overlapping patches. Each patch is then flattened and linearly projected into a lower-dimensional embedding space, forming tokens. Suppose the image is divided into \( N \) patches, each of size \( P \times P \), resulting in a sequence of \( N \) tokens. This can be represented as:
\begin{equation}
    \mathcal{E}(X) = [t_1, t_2, \dots, t_N],
\end{equation}
where \( t_i \in \mathcal{R}^{E} \) is the embedding of the \( i \)-th patch, and \( E \) is the embedding dimension.
To enable classification, a special trainable token, known as the \emph{class token} (\text{CLS} token), is prepended to the sequence of patch embeddings. Additionally, since transformers lack an inherent sense of order, positional encodings are added to each token to retain spatial information. 
%
The transformer layers \( \mathcal{F}(\cdot) \) process this sequence of tokens through a series of operations involving multi-head self-attention and feed-forward neural networks. A generic transformer block at layer \( l \) transforms each token from layer \( l-1 \) via:
\begin{equation}
    t_l = \mathcal{F}^{l}(t_{l-1}),
\end{equation}
where \( t_{l-1} \) denotes the token embedding at layer \( l-1 \), \( t_l \) the updated token embedding, and \( \mathcal{F} \) represents a standard transformer encoder block with multi-head attention and a feed-forward MLP.
The self-attention operation has a quadratic complexity with respect to the sequence length, i.e. it has a cost of \( \mathcal{O}(N^2) \), where $N$ is the number of tokens. This quadratic cost can be significant for devices with limited resources. Efficient implementation techniques or approximations are often required to make ViTs faster and more memory efficient for inference on downstream tasks, especially in low resource settings ~\cite{meng2022adavit, yin2022adaptivevit, bolya2022tokenmerging}.  

The \text{CLS} token, which aggregates information from all patches, is extracted after the final transformer layer and passed to the classification head \( \mathcal{C}(\cdot) \). This head typically consists of a linear layer followed by a softmax function to produce the final classification output. We show an overview of the Vision Transformer architecture adapted to our method in \Cref{fig:method}.
\subsection{Layer Contributions during Fine-tuning}
We analyze the transformer architecture from the perspective of the residual stream, as described by \citet{elhage2021mathematical}. In this framework, the set of tokens flows through the model, with token embeddings being updated via vector additions from the attention and feed-forward blocks in each layer. This perspective allows us to isolate and examine the individual contributions that each layer adds to the residual stream.

Multiple studies \citep{samragh2023weightsub, zhang2022layers, unreasonable} have demonstrated that not all layers contribute equally to the updates in the residual stream. This phenomenon is particularly evident in pre-trained models, where some layers function almost like identity mappings, providing minimal updates to the token embeddings in the residual stream.

\begin{figure}[t]
    \centering
    \hspace*{-0.5cm} 
    \includegraphics[width=1.05\linewidth]{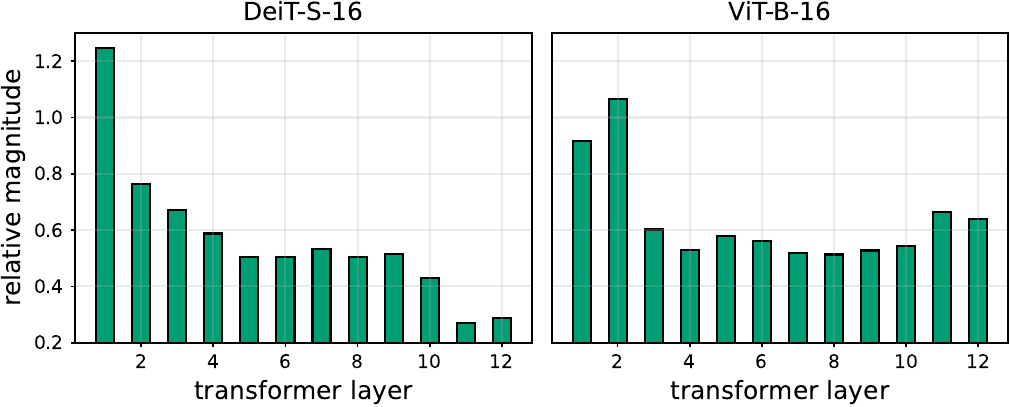}
    \caption{Relative magnitude $\frac{|f(x)|}{|f(x)+x|}$ for each transformer layer in pre-trained \deits (left) and  \vit (right). Layers with low relative magnitudes (final ones for \deits and middle ones for  \vit) provide a minimal contribution to the residual token stream, working as identity functions.}
    \label{fig:rel_mag}
\end{figure}

To visually illustrate this behavior, we define the relative magnitude of a transformer layer as the ratio of the non-residual block’s contribution to the overall layer output. Specifically, given the layer input 
$x$ and the layer's contribution $f(x)$ , we follow \citet{samragh2023weightsub} and plot the relative magnitude $\frac{|f(x)|}{|f(x)+x|}$ for each transformer layer. Small values indicate that the layer is leaving the input largely unchanged. We present an example of this for  \vit and  \deits in \Cref{fig:rel_mag}.

From an efficiency standpoint, performing computations in layers that do not significantly alter the input embeddings is a waste of computational resources, especially when those layers are being trained.
This insight has been applied in various ways, such as distilling large language models into smaller ones \cite{unreasonable} or initializing smaller transformers for fine-tuning \cite{samragh2023weightsub}. More recently, \citet{bsr} identified the most important layers in a ViT and focused on fine-tuning only those layers. While these strategies produce competitive results, they come with the drawback of requiring extensive experimentation to identify the optimal combination of layers to train. In contrast, we introduce a method that automatically learns which layers are most important, eliminating the need for such experimentation.
\section{Adaptive Layer Selective Fine-tuning}
\begin{figure*}[t]
\centering
  \includegraphics*[width=0.9\textwidth]{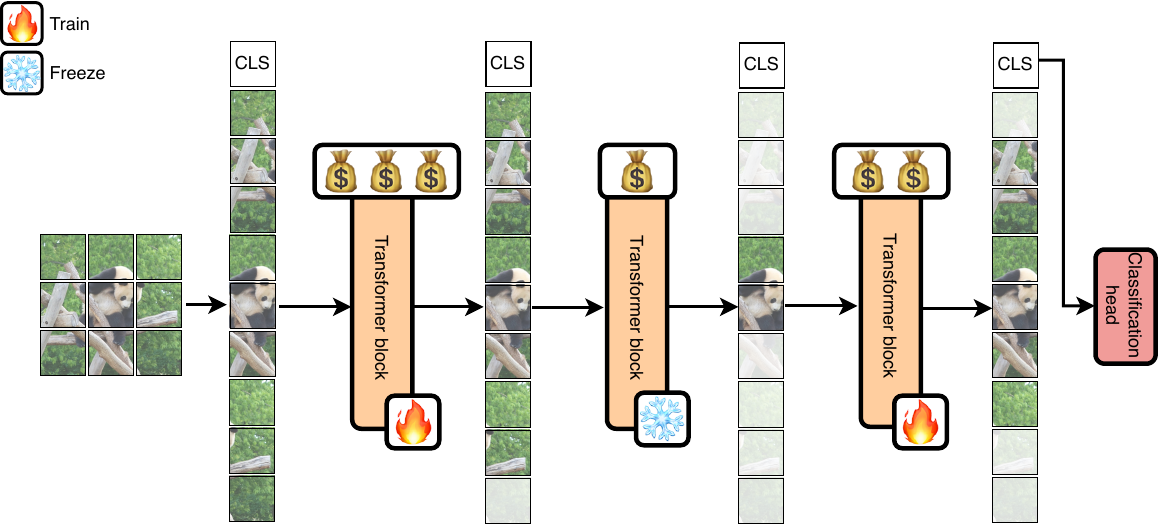}
  \caption{At each fine-tuning iteration, each layer is assigned a compute budget. Based on the budget, we allow more (high budget) or fewer (low budget) tokens to flow through the layer - greyed out tokens are excluded from computation. Additionally, we freeze layers with the lowest budgets to save computing resources and memory.}
  \label{fig:method}
\end{figure*}
%
Because different layers affect the final prediction in different ways \cite{dodge2020fine, samragh2023weightsub}, pre-determining which layers to fine-tune can result in sub-optimal outcomes. Previous approaches have attempted to identify the most critical layers through extensive search methods \cite{samragh2023weightsub, bsr, unreasonable}. However, these methods are highly dependent on the specific dataset and model being used, leading to inconsistencies when applied to different data distributions or model scales. For example, an exhaustive search might determine that layers 4, 5, and 6 are the most important when fine-tuning on the \flowers dataset, leading to the decision to freeze the other layers. Yet, this configuration may not be effective for a different dataset, necessitating another round of exhaustive searching. Similarly, the important layers in  \vit may differ from those in \deitt, requiring a separate search for each model.

To overcome these challenges, we propose a simple strategy that \emph{adaptively} estimates the importance of each layer during fine-tuning, leading to improvements in memory usage, FLOPs, and wall-clock training time. This approach is especially valuable in low-resource scenarios, where it can be integrated requiring minimal modifications to existing training pipelines. In such scenarios where resources are limited, we argue that not all layers should be trained with the same computational effort; some layers should receive a reduced computational budget and therefore be trained with fewer resources.

To optimize resource allocation, we focus on two key parameters: the number of tokens processed by each layer and which layers are actively trained. Adjusting these parameters directly influences computational load and memory consumption. For the first parameter, we follow \citet{meng2022adavit} and \citet{rao2021dynamicvit} by reducing the number of tokens processed, selectively removing redundant ones during the forward pass. For the second parameter, we freeze less critical layers, saving memory and preventing unnecessary updates to their weights during the backward pass (\Cref{fig:first_page}).
We highlight that the proposed method can be integrated into existing fine-tuning frameworks with minimal overhead. We provide an overview of \ours in \Cref{fig:method}. To implement the budget allocation, we first need a reliable method to estimate the importance of each layer. 
\subsection{Estimating the importance of each layer}
In the following, we explain how we assign a budget to each transformer layer. We assume that each fine-tuning step comprises a forward and a backward pass. We use $i$ to indicate the current step and $ \text{CLS}^{i}_{l}$ to indicate the embedding of the class token at layer $l$ for step $i$.
Given a fine-tuning step $i$, we estimate the importance of each layer $l$ and assign it a training budget $b^{i}_{l} \in (0,1)$, representing the computational resources we can afford to spend on that layer. The budget should be high if the layer's contribution to the final prediction is high, low otherwise.  Notably, the budget must be computed \textit{adaptively at each step}. 

During the forward pass, the class token aggregates information from all other tokens and it is passed through the classification head. This token is crucial for capturing rich semantic information about the input, making it a reliable indicator for evaluating each layer's contribution to the final prediction. The class token's role in capturing essential features for downstream tasks has already been investigated in previous works, such as \citet{raghu2021do}, which studied the correlation between the  \text{CLS} token representation and each layer's contribution to the final prediction.
This correlation can be intuitively understood by noting that the class token, being the only one that is passed through the final classifier, must capture information about the whole input image \cite{liang2022evit, raghu2021do}. 
As a result, layers that contribute less to updating the class token are less critical, and we assign them lower compute budgets without impacting overall performance.

We now provide a more detailed explanation of how we estimate the appropriate budget for each layer based on this observation. At fine-tuning step $i$, layer $l$ updates the class token according to:
\begin{equation}
     \text{CLS}^{i}_l = \mathcal{F}^{l} ( \text{CLS}^{i}_{l-1}) 
\end{equation}
where $\text{CLS}^{i}_{l-1}$ is the class token representation coming from the previous layer, $\mathcal{F}^{l}$ represents the function applied by layer $l$, and $\text{CLS}^{i}_l$ is the updated class token representation.
To capture the variations in the class token embedding, we define the class token delta at layer $l$ and training step $i$ as:
\begin{equation}
    \Delta^{i}_{l} = {( \text{CLS}^{i}_l -  \text{CLS}^{i}_{l-1})}^2
\end{equation}
The class token delta quantifies the magnitude of the update performed by layer $l$ on the class token's residual stream. Finally, we use the variation in class token deltas across iterations to compute the training budget. At the next training iteration $i+1$, we compute:
\begin{equation}
    b^{i+1}_{l} = b^{i}_l + ( \Delta^{i+1}_{l} - \Delta^{i}_{l} ) \times \alpha
\end{equation}
Here, $b^{i}_{l}$ represents the budget from the previous training iteration, and $\alpha$ is the budget learning rate, which we initialize to match the fine-tuning learning rate. We always initialize the budget to $1$ for each layer, and clip it during fine-tuning to have a value between $0$ and $1$.
By assigning a budget to each layer, we obtain a distribution of budgets $\mathcal{D}^{i}_{L}$, where $L$ is the number of layers and $i$ is the current training step. We provide an example of how compute budgets vary during fine-tuning in \Cref{fig:deit_s_food_distributions}. 
We then leverage this budget distribution to allocate resources appropriately across layers.
\subsection{Applying the budget}
Based on the distribution of budgets $\mathcal{D}^{i}_{L}$, we allocate our memory and FLOPs resources leveraging two degrees of freedom: the number of processed tokens and the trainable layers. In the two following paragraphs, we explain how we select which tokens to discard and which layers to freeze.
\paragraph{Adaptive Token selection} \label{sec:methods}
As pointed out in several works \cite{jain2024mixturenestedexpertsadaptive, meng2022adavit, bolya2022tokenmerging} not all tokens carry equally valuable information for the task at hand. Given an input sequence $T \in \mathcal{R}^{N \times E}$, where $N$ is the sequence length and $E$ is the embedding dimension, we only allow $b \cdot N$ tokens to flow to the next layer. Here, $b \in [0,1]$, meaning that this approach selects $b\%$ of the tokens from the input sequence. 
%
%
To determine which tokens to discard, we adopt a strategy similar to that of \citet{liang2022evit}. We rank the tokens 
based on their attention scores from the \text{CLS} token and retain only the top $b \cdot N$ tokens.

\begin{figure}[t]
    \centering
    \begin{subfigure}[b]{0.32\linewidth}
        \centering
        \includegraphics[width=\linewidth]{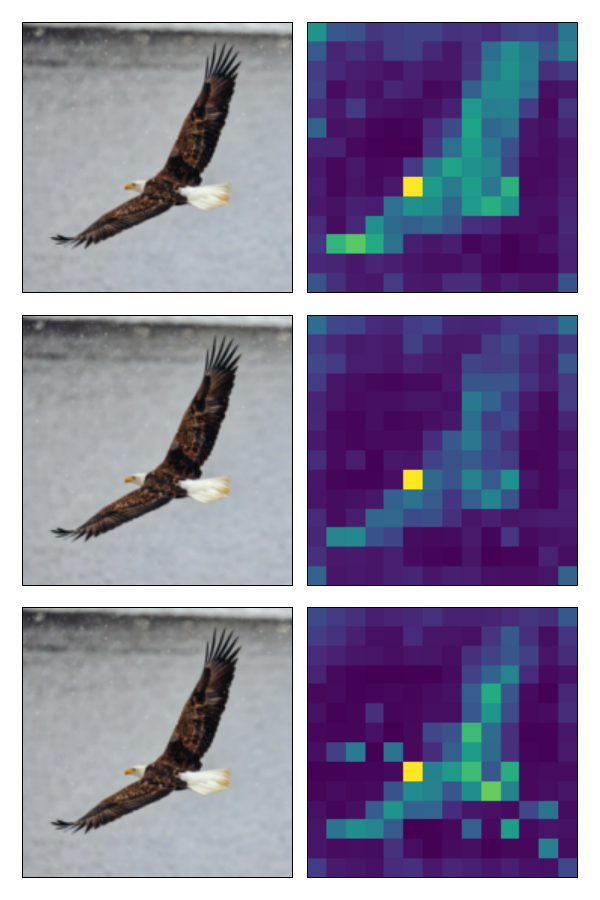}
    \end{subfigure}
    \hfill
    \begin{subfigure}[b]{0.32\linewidth}
        \centering
        \includegraphics[width=\linewidth]{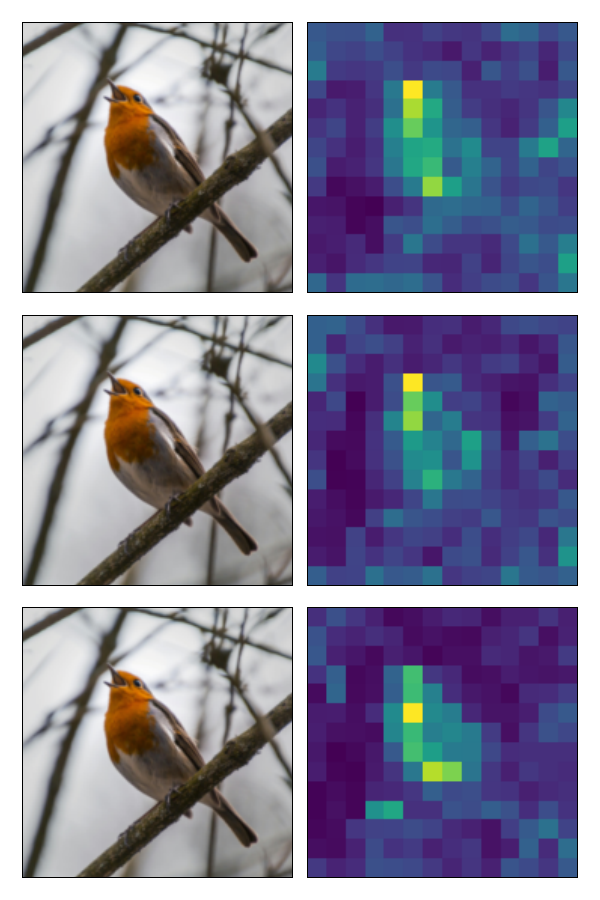}
    \end{subfigure}
    \hfill
    \begin{subfigure}[b]{0.32\linewidth}
        \centering
        \includegraphics[width=\linewidth]{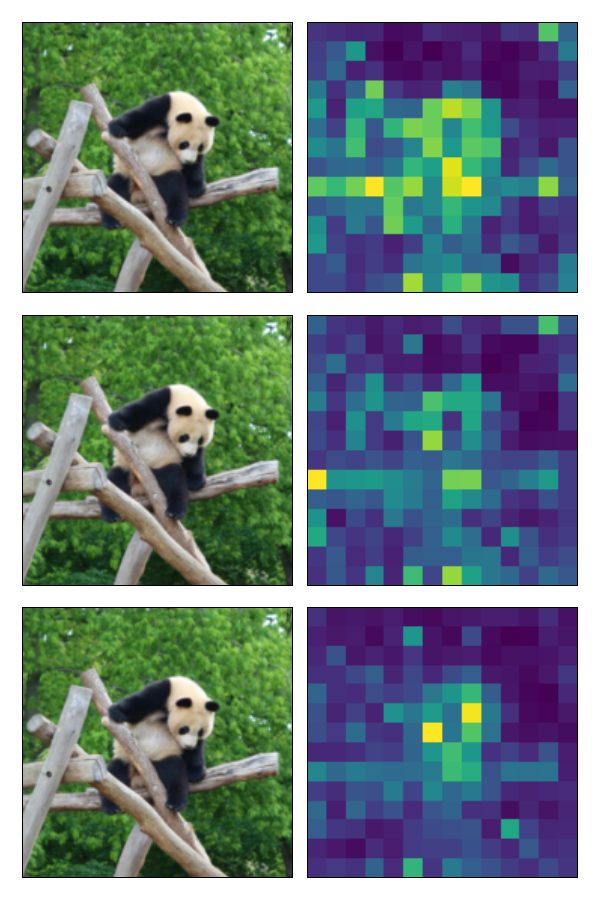}
    \end{subfigure}
    \caption{Attention of  \text{CLS} token for different patches at layer 2,4,6 of \deits \cite{touvron2019deit}. Brighter patches have higher attention.  \text{CLS} token's attention captures semantically important patches. 
    }
    \label{fig:cls_token}
\end{figure}

This approach leverages the fact that tokens with higher attention scores are more influential in determining the class prediction, as we show in \Cref{fig:cls_token}, where we plot some attention maps with respect to the class token for \deits. By retaining only the most important tokens - those that contribute significantly to the class token's representation - we reduce computational overhead and memory usage while preserving the most relevant information for classification.
Initially, we allowed unattended tokens to proceed to the next layer. However, analysis of attention score distributions revealed that tokens excluded at one layer remained excluded in subsequent layers. Consequently, we opted to discard the least attended tokens, preventing their progression to the next layer. This approach further reduced batch size and memory consumption without performance loss.
\paragraph{Adaptive Layer Freezing}
In addition to token selection, we also optimize resource allocation by freezing certain layers. Given a budget $b$, we freeze the less critical layers to conserve memory, which is crucial for efficient on-device training. By freezing these layers, we reduce the number of trainable parameters, thereby lowering memory consumption and improving computational efficiency. When a layer is frozen, we save the memory which is necessary to store the activations for the backpropagation.
After determining the budget distribution $\mathcal{D}^{i}_{L}$, we sample layers $\mathbf{x} \sim \mathcal{D}^{i}_{L}$ without replacement, ensuring that the same layer is not selected twice. We then train only the $K$ layers with the highest budget allocation, and freeze the rest.
The dual approach of token selection and layer freezing enables us to manage resources effectively and optimize the performance of our model within the constraints of the available device memory, by leveraging two different degrees of liberty. 
\section{Experimental Setup}
We fine-tune our models 
on \flowers \citeauthor{flowers102}, \cifar \citeauthor{cifar100} and the more challenging \food \citeauthor{bossard14} dataset. We test a larger pre-trained Vision Transformer \vit \cite{dosovitskiy2021vit} along with the \deits and \deitt \cite{touvron2019deit} models, that are smaller and more parameter efficient, thus more likely to be used in resource-constrained scenarios. We show the tested model architectures in \Cref{tab:models}.  We download pre-trained weights from timm-models \cite{rw2019timm}.
In all the runs, we set $K$ (number of trainable layers) to 9, but we show results for other values in \Cref{sec:appc}. 
We use mixed precision training \cite{micikevicius2018mixed} for all our experiments to keep memory usage as small as possible and simulate a real-world on-device training. 
During training, we keep track of FLOPs (Floating Point Operations), memory load and wall-clock time.  While memory and time are important metrics to assess practical performance and resource utilization, they can be partially influenced by hardware. Therefore, we also use FLOPs to provide a hardware-independent measure of computational complexity, enabling consistent cross-system comparisons.  
For reproducibility, we provide detailed descriptions of our experimental setup, including code, hyperparameters, and training procedures in \Cref{sec:appe}.
\begin{table}[t]
    \centering
    \scalebox{0.8}{
        \begin{tabular}{l|ccccc}
        \toprule
        \textbf{Model} & \textbf{embedding}  & \textbf{\#heads} & \textbf{\#layers} & \textbf{\#params}   \\
        \midrule
         \deitt  & 192 & 3 & 12 & 5M \\
         \deits  & 384 & 6 & 12 & 22M\\
         \vit   & 768 & 12 & 12 & 86M \\
        \bottomrule
        \end{tabular}
        }
    \caption{Variants of tested ViT architectures.}
    \label{tab:models}
\end{table}
\section{Results \& Analysis}
\begin{figure}[b]
    \centering
    \includegraphics[width=0.8\linewidth]{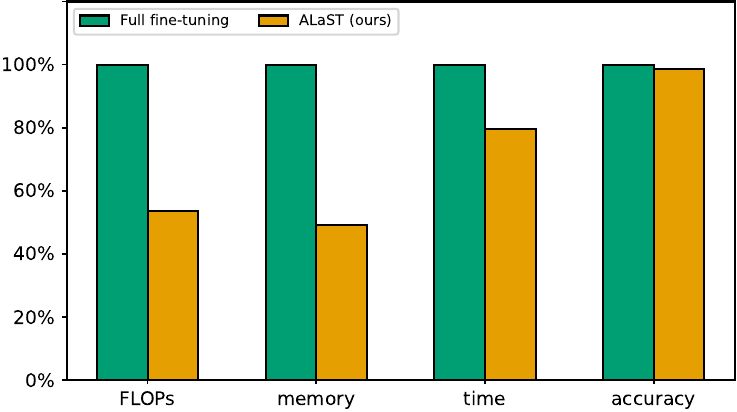}
    \caption{Normalized improvement when fine-tuning with \ours with respect to full fine-tuning for FLOPs, Memory, wall-clock time and accuracy. On average, we achieve similar accuracy, with $60 \%$ FLOPs, $50 \%$ memory and $80 \%$ time.  We average the results on all the considered datasets for  \deits. }
    \label{fig:overall_stats}
\end{figure}
\begin{figure*}[t]
    \centering
    \begin{subfigure}[b]{0.3\textwidth}
        \centering
        \includegraphics[width=\textwidth]{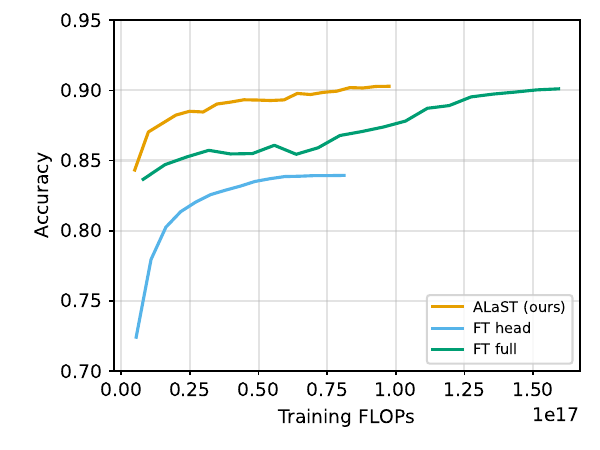}
    \end{subfigure}
    \hfill
    \begin{subfigure}[b]{0.3\textwidth}
        \centering
        \includegraphics[width=\textwidth]{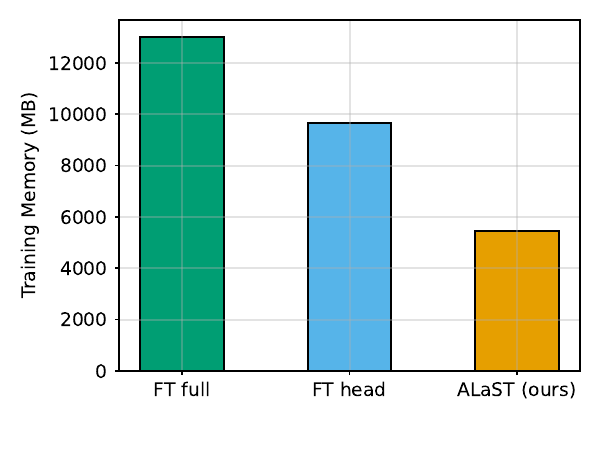}
    \end{subfigure}
    \hfill
    \begin{subfigure}[b]{0.3\textwidth}
        \centering
        \includegraphics[width=\textwidth]{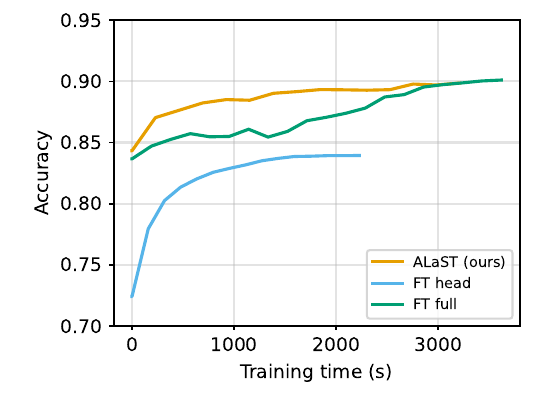}
    \end{subfigure}
    \caption{Comparison for  \vit on a fixed data budget of 20 epochs for \food. Our method converges with fewer FLOPs (left), lower memory (center)  and shorter time (right), with respect to fine-tuning all layers (FT full) and only head (FT head).}
    \label{fig:ours_first_comp}
\end{figure*}
We evaluate our method against a diverse array of baselines, including both traditional fine-tuning approaches and state-of-the-art techniques, including Low-Rank Adaptation (LoRA) \cite{hu2022lora}, Token merging \cite{bolya2022tokenmerging} and Block Selective Reprogramming \cite{bsr}. 
Compared to standard fine-tuning procedures, \ours achieves optimal performance at a fraction of the cost. While fine-tuning all layers typically results in high final accuracy, it comes at the expense of significantly more expensive training. In \Cref{fig:overall_stats}, we demonstrate the normalized gains our method achieves compared to full fine-tuning. With \ours, we observe only a minimal reduction in accuracy while using just $60\%$ of the FLOPs, $50\%$ of the memory, and $80\%$ of the fine-tuning wall-clock time relative to the full fine-tuning baseline,  without adding any additional parameters.
An alternative to full fine-tuning is training only the classification head, which is computationally and memory efficient since only the final linear layer is updated via back-propagation. However, this approach typically results in substantially lower final accuracy. \ours strikes a balance between these two extremes, maintaining low memory usage and training FLOPs, while losing only minimal accuracy compared to full fine-tuning. A more detailed comparison can be found in \Cref{fig:ours_first_comp}.
In low-resource scenarios, it is critical not only to reduce the total resources consumed during fine-tuning but also to manage the distribution of these resources throughout the training process. Often, training until full convergence is not feasible, making the initial "speed" of convergence particularly important. In other words, a method that requires fewer FLOPs but converges slowly is impractical for such constrained environments. As shown in \Cref{fig:ours_first_comp}, thanks to its dynamic compute budget allocation, \ours converges to higher accuracy in less time and with fewer FLOPs, making it well-suited for resource-constrained scenarios. 
%
\paragraph{Comparison to other methods}
In \Cref{table:food101} and \Cref{table:cifar100}  we provide more detailed results on \food and \cifar, and report results on \flowers in \Cref{sec:appa} to conserve space. For each method, we track training FLOPs, peak training memory, and wall-clock time. We compare our approach to six different fine-tuning baselines. Token Merging (ToMe) \cite{bolya2022tokenmerging} merges $r$ tokens at each transformer layer, using an optimal $r$ as determined by hyperparameter search. For Block Selective Reprogramming (BSR), we follow the methodology proposed in \cite{bsr} and select the three most influential blocks, while simultaneously discarding tokens. Another baseline involves identifying the three most influential blocks through exploration and training only those blocks (FT top-3). Finally, we also include a comparison with the popular LoRA method \cite{hu2022lora}.
A key distinction of our method is that, unlike ToMe, FT top-3, and BSR, it \textit{adaptively} learns which layers are most important, eliminating the need for extensive experimentation to identify the crucial layers in advance. Compared to LoRA, our method does not add any parameters to the model.

\begin{figure}[h!]
    \centering
    \begin{subfigure}[b]{0.49\linewidth}
        \centering
        \includegraphics[width=\textwidth]{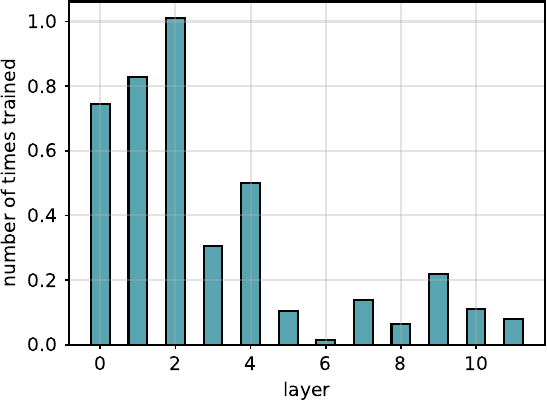}
    \end{subfigure}
    \hfill
    \begin{subfigure}[b]{0.49\linewidth}
        \centering
        \includegraphics*[width=\textwidth]{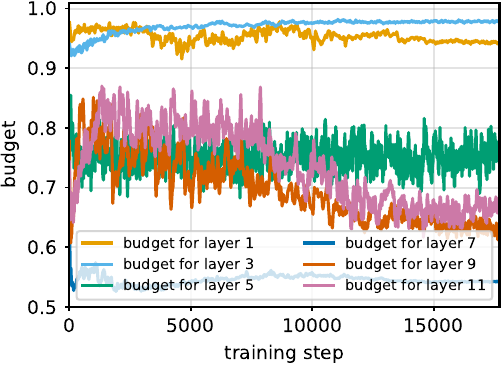}
    \end{subfigure}
    \centering
    \begin{subfigure}[b]{0.49\linewidth}
        \centering
        \includegraphics[width=\textwidth]{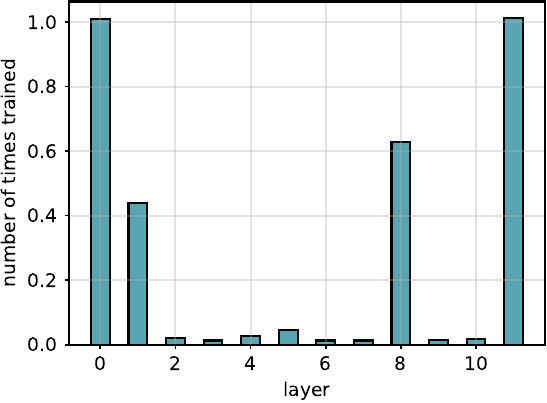}
    \end{subfigure}
    \hfill
    \begin{subfigure}[b]{0.49\linewidth}
        \centering
        \includegraphics*[width=\textwidth]{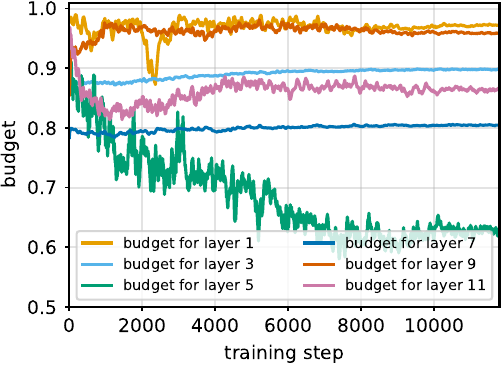}
    \end{subfigure}
    \caption{Compute budget allocation during fine-tuning of \deits (top) and \deitt (bottom) on \food. Left: absolute training frequency per layer. Right: budget variation across training. Additional plots for different models and datasets are in \Cref{sec:appb}}
    \label{fig:deit_s_food_distributions}
\end{figure}

While ToMe is effective at minimizing memory consumption, it struggles to achieve high accuracy, particularly with larger models. This limitation likely stems from the fixed $r$ schedule, which is predetermined and not adjusted during training, causing some layers to "starve" at different stages of fine-tuning. In contrast, \ours dynamically allocates computational resources to the most critical layers, leading to a more efficient use of the computational budget.
BSR, on the other hand, preselects which layers to freeze and discards tokens accordingly. This preselection results in a suboptimal combination of frozen layers, leading to lower accuracy and higher memory usage during fine-tuning compared to our method.
%
%
\begin{table*}[t]
\caption{Fine-tuning  \deitt,  \deits and  \vit on \food. Proposed method results and best results in bold.}
\label{table:food101}
\centering
\scalebox{0.7}
{
\begin{tabular}{l | c c c c | c c c c | c c c c}
\hline
\toprule
\textbf{Model} & \multicolumn{4}{c|}{\textbf{Deit-t-16}} & \multicolumn{4}{c|}{\textbf{Deit-s-16}} & \multicolumn{4}{c}{\textbf{ \vit}}  \\[0.2cm]
\textbf{} & \parbox{1.2cm}{\centering \textbf{Compute} \\  (PFLOPs)} & \parbox{1.2cm}{\centering \textbf{Memory} \\  (MB)} & \parbox{1.2cm}{\centering \textbf{Time} \\  (min)} & \textbf{Accuracy} & \parbox{1.2cm}{\centering \textbf{Compute} \\  (PFLOPs)} & \parbox{1.2cm}{\centering \textbf{Memory} \\ (MB)} & \parbox{1.2cm}{\centering \textbf{Time} \\  (min)} & \textbf{Accuracy} & \parbox{1.2cm}{\centering \textbf{Compute} \\  (PFLOPs)} & \parbox{1.2cm}{\centering \textbf{Memory} \\ (MB)} & \parbox{1.2cm}{\centering \textbf{Time} \\  (min)} & \parbox{1.2cm}{\centering \textbf{Accuracy} \\  (top-1)} \\
\midrule
FT head & \textbf{7.5} & \textbf{2389} & \textbf{13.44} & 0.59 & 10.5 & 4751 & \textbf{18.36} & 0.70 & \textbf{80} & 9657 & \textbf{35.47} & 0.83  \\ 
FT full  & 11.2 & 3267 & 24.13 & \textbf{0.83}  & 30.2 & 7186 & 33.08 & \textbf{0.86} & 150 & 13012 & 64.37 & \textbf{0.90} \\ 
FT top-3 & 9.6 & 2525 & 15.17 & 0.8 & 22.8 & 5202 & 20.40 & 0.84 & 120 & 7096 & 39.45 & 0.90  \\ 
ToME \citenump{bolya2022tokenmerging} & 7.5 & 2983 & 18.06 & 0.82 & \textbf{10.0} & \textbf{2134} & 25.05 & 0.80 & 153 & \textbf{4493} & 48.42 & \textbf{0.88}  \\   
BSR \citenump{bsr} & 8.5 & 2500 & 14.30 & 0.80 & 12.3 & 3520 & 19.38 & 0.83 & 117 & 6900 & 37.46 & 0.88  \\ 
LoRA \citenump{hu2022lora} & 8.2 & 2620 & 26.44 & 0.78 & 22.5 & 5060 & 35.12 & 0.83 & 109 & 11823 & 48.35 & 0.89   \\ 
\ours (ours) & \textbf{5.1} & \textbf{1724} & \textbf{18.0} & \textbf{0.81} & \textbf{14.5} & \textbf{4190} & \textbf{25.05} & \textbf{0.86} & \textbf{107} & \textbf{5451} & \textbf{48.42} & \textbf{0.90}  \\ 
\bottomrule
\end{tabular}
}
\end{table*}
%
\begin{table*}[t]
    \caption{Fine-tuning  \deitt,  \deits and  \vit on \cifar. Proposed method results and best results in bold.}
    \label{table:cifar100}
    \centering
    \scalebox{0.7}
    {
    \begin{tabular}{l | c c c c | c c c c | c c c c}
    \hline
    \toprule
    \textbf{Model} & \multicolumn{4}{c|}{\textbf{Deit-t-16}} & \multicolumn{4}{c|}{\textbf{Deit-s-16}} & \multicolumn{4}{c}{\textbf{ \vit}}  \\[0.2cm]
    \textbf{} & \parbox{1.2cm}{\centering \textbf{Compute} \\  (PFLOPs)} & \parbox{1.2cm}{\centering \textbf{Memory} \\  (MB)} & \parbox{1.2cm}{\centering \textbf{Time} \\  (min)} & \textbf{Accuracy} & \parbox{1.2cm}{\centering \textbf{Compute} \\  (PFLOPs)} & \parbox{1.2cm}{\centering \textbf{Memory} \\ (MB)} & \parbox{1.2cm}{\centering \textbf{Time} \\  (min)} & \textbf{Accuracy} & \parbox{1.2cm}{\centering \textbf{Compute} \\  (PFLOPs)} & \parbox{1.2cm}{\centering \textbf{Memory} \\ (MB)} & \parbox{1.2cm}{\centering \textbf{Time} \\  (min)} & \textbf{Accuracy} \\
\midrule
FT head     & \textbf{5.0} & \textbf{2389} &   \textbf{8.18} & 0.66 & \textbf{18.4} & 4751 & \textbf{11.34} & 0.75 & \textbf{70} & 9657 & \textbf{22.40} & 0.84 \\ 
FT full     & 7.8 & 3267 &   14.14 & \textbf{0.84} & 27.1  & 7186  &20.14  & \textbf{0.88} & 90   & 13012  & 40.34  & 0.90    \\ 
FT top-3    & 6.3 & 2525 &   9.13 & 0.82  &  21.0  & 5202 & 12.51 &  0.87 & 80 & 7096  & 24.11   & \textbf{0.92}   \\ 
ToME \citenump{bolya2022tokenmerging} & 7.1 & 2983 & 11.32 & \textbf{0.84} & 26.7  & \textbf{3122} & 35.23 & 0.88 & 75  & \textbf{4493}  & 30.29  & 0.86 \\ 
BSR \citenump{bsr} & 6.0 & 3052 & 8.45 & 0.82 & 13.0 & 3500 & 12.13  &0.87  &  79 & 6900  & 23.30 & 0.91   \\ 
LoRA \citenump{hu2022lora}  & 5.5 & 2643 & 15.10 & 0.81 & 22.3 & 5060 & 32 & 0.85 & 80  & 11823 & 32.10  & 0.90  \\ 
\ours (ours)  & \textbf{5.0} & \textbf{2397} & \textbf{11.32} & \textbf{0.83} & \textbf{15.5} & \textbf{3613}  & \textbf{15.05} & \textbf{0.87} & \textbf{70} & \textbf{5341}  & \textbf{29.30} & \textbf{0.90} \\ 
\bottomrule
\end{tabular}
}
\end{table*}
We observe that LoRA's accuracy degrades significantly with smaller  \deitt and  \deits models. This is likely due to the reduced representational capacity of smaller models, that makes it hard to adjust to a new distribution only leveraging LoRA's additional parameters, leading to relatively lower performance improvements.  On larger  \vit, LoRA seems to achieve a similar performance to our method at the cost of a higher memory load needed to store all the activations. Because our method is orthogonal to PEFT strategies in general, we also experiment with integrating it with LoRA in \Cref{sec:appd}.
\paragraph{Layer Budgets across Fine-tuning}
Budget assignement is dynamic and happens at each fine-tuning iteration. For  \deits and  \vit, higher compute budgets are usually allocated to the first layers already in the initial part of the fine-tuning, as we show in \Cref{fig:deit_s_food_distributions}. Interestingly, for the first two layers, the method allocates high budgets across the entire fine-tuning, while for central layers the budget usually decreases significantly or exhibits a higher variance. For  \deitt, we observe a different pattern, where first and last layers are allocated higher budgets, while central ones are often neglected. We show more plots and results analyzing budget allocation in \Cref{sec:appb}.
\section{Related Works}
\subsection{Parameter Efficient Fine-Tuning}
In the last years, the field of parameter efficient fine-tuning (PEFT) has seen increased interest. One of the first approaches in PEFT was the use of adapters, \citeauthor{adapters}, that insert and fine-tune small neural network modules into each layer of a pre-trained model. Prefix-tuning \citeauthor{li-liang-2021-prefix} and prompt-tuning \citeauthor{prompt-tuning} learn soft prompts that are added to the input tokens. Perhaps the most popular PEFT approach is LoRA (Low-Rank Adaptation), proposed in \cite{hu2022lora}, that reduces the number of additional parameters that need to be fine-tuned by decomposing the weight updates into a product of two low-rank matrices. Most of the PEFT methods were introduced for fine-tuning large language models, and subsequently applied to computer vision with varying degrees of success \cite{xin2024parameterefficientfinetuningpretrainedvision}. Typically, these approaches emphasize optimizing the number of additional parameters rather than FLOPs or training time. However, parameters memory is only a fraction of the memory used during training. Unlike PEFT, \ours does not add any parameters, and also reduces FLOPs and memory. Additionally, \ours can be combined with LoRA, as we show in \Cref{sec:appd}.
\subsection{On-device Training}
Our work is perhaps most similar to those papers that optimize resources for on-device training. Among these, \cite{tinytl} was the first to highlight that memory is a major bottleneck in CNN training and devise a way to reduce memory load. Along the same line, some approaches proposed to recompute activations for a subset of the layers to reduce memory consumption \cite{chen2016trainingdeepnetssublinear, btt_eff}. While this strategy can be useful for low memory budgets, it sacrifices the compute (FLOPs), and is not suitable for devices with limited computational resources. Recently, \citet{bsr, samragh2023weightsub} explored smart initialization by identifying the best subset of layers to fine-tune. \citet{bsr} identifies the most important layers and discards redundant tokens during fine-tuning. \citet{samragh2023weightsub} uses the relative magnitude of outputs to estimate the importance of different blocks and initialize a smaller model from a larger one. Unlike these methods, \ours learns the importance of individual layers during fine-tuning and therefore does not need extensive research of which layers to train in advance.
\subsection{Efficient Transformers}
A substantial body of research has focused on enhancing the efficiency of ViTs, particularly during the inference phase. 
For example, \cite{ rao2021dynamicvit, meng2022adavit} proposed  token halting strategies that select the most important tokens at runtime and speed up inference. \citet{touvron2019deit} employed distillation from CNNs to maintain model accuracy while reducing the number of parameters and FLOPs. \citet{lin2022fqvit} introduced a quantized version of the Vision Transformer to decrease model inference complexity. More recently, \citet{acm} proposed to control the computational load by activating sub-modules of the transformer, based on input difficulty. In contrast to these methods, we do not focus on inference, but address the challenge of fine-tuning available foundation models when resources are limited. 
\section{Conclusions}
We introduced \ours, a simple and effective method to fine-tune ViTs in low-resource scenarios, saving computational budget, memory load and training time, with minimal modifications to the training pipeline. Although we test \ours on ViTs, its principles are generalizable to other transformer-based architectures, which we plan to explore in future work.
\section*{Acknowledgements}
This work has been supported by the SNS JU project 6G-GOALS under the EU’s Horizon program Grant Agreement No 101139232, by Sapienza grant RG123188B3EF6A80 (CENTS), and by European Union under the Italian National Recovery and Resilience Plan of NextGenerationEU, partnership on Telecommunications of the Future (PE00000001 - program RESTART)
\bibliography{aaai25}

\clearpage

\appendix

\section{Additional Results} 
\label{sec:appa}

We show additional results on the \flowers dataset in table \Cref{table:flowers102}. \flowers is the smallest dataset that we use for our experiments.

\section{Number of trainable blocks} 
\label{sec:appc}
\begin{figure}[b]
    \centering
    \includegraphics[width=0.8\linewidth]{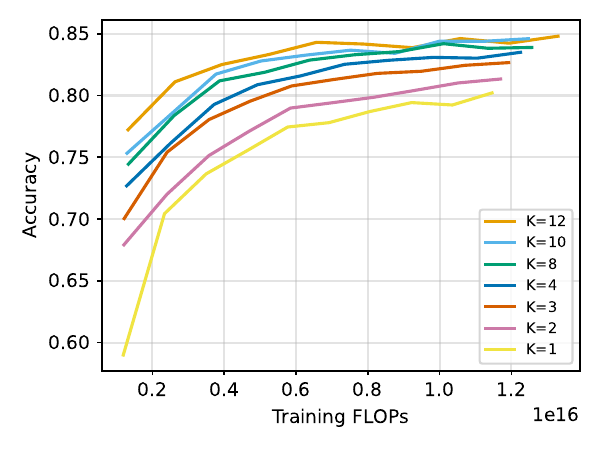}
    \caption{Accuracy vs FLOPs trade-off during fine-tuning, with different number of trainable layers for \deits on \food.}
    \label{fig:k_ablation}
\end{figure}

At each iteration, we select $K$ transformer layers for fine-tuning. We showed experiments with $K=9$. Here we report results for different values of $K$. In \Cref{fig:k_ablation} we show the accuracy and FLOPs trade-off for different values of $K$. We see that incresing the number of trainable layers leads to better performance in general. The improvement becomes smaller when $K$ is greater than $8$. We report detailed results in \Cref{table:k_ablation}. 

\begin{table}[b]
    \caption{Different values for the number of trainable blocks at each iteration, $K$.}
    \label{table:k_ablation}
    \centering
    \scalebox{0.8}
    {
    \begin{tabular}{l | c c c c}
    \hline
    \toprule
        \textbf{$K$} & \parbox{1.5cm}{\centering \textbf{Compute} \\ (PFLOPs)} & \parbox{1.5cm}{\centering \textbf{Memory} \\ (MB)} & \parbox{1.5cm}{\centering \textbf{Time} \\ (min)} & \textbf{Accuracy} \\
        \midrule
        1   &  11.6 & 3125  & 20 & 0.81 \\ 
        2   &  11.9 & 3211  & 21.30 & 0.83  \\ 
        3   &  12.0 & 3297  & 22.20 & 0.84 \\
        4   &  12.4 & 3383  & 23.50 & 0.85 \\
        8   &  13.0 & 3888  & 23.10 & 0.86 \\
        10  &  15.5 & 4000  & 24.10 & 0.86 \\
        12  &  15.8 & 4137  & 25.10 & 0.86 \\
        \bottomrule
    \end{tabular}
    }
\end{table}
\section{Combination with PEFT} \label{sec:appd}
As we showed in \Cref{table:cifar100} and \Cref{table:food101}, LoRA performs rather poorly when applied to smaller models, while \ours converges faster and to higher accuracy.  On larger  \vit, on the other hand, LoRA achieves similar accuracy to our method, at a higher memory cost to store the activations for all layers. Because the two methods are orthogonal, we integrate them and test LoRA + \ours on   \vit. 

In order to combine LoRA with \ours, we apply the budget schedule to LoRA's additional parameters and keep the transformer layers frozen, except for the classfication head. We report results for \cifar and \food in \Cref{table:LoRA_combine_cifar} and \Cref{table:LoRA_combine_food}, respectively. By intergrating LoRA with \ours, we further reduce the memory footprint, while still achieving accuracy comparable to baseline.
 
\begin{table}[t]
    \caption{Combining LoRA + \ours for fine-tuning  \vit on \cifar dataset.}
    \label{table:LoRA_combine_cifar}
    \centering
    \scalebox{0.8}
    {
    \begin{tabular}{l | c c c c}
    \hline
    \toprule
        \textbf{Method} & \parbox{1.5cm}{\centering \textbf{Compute} \\ (PFLOPs)} & \parbox{1.5cm}{\centering \textbf{Memory} \\ (MB)} & \parbox{1.5cm}{\centering \textbf{Time} \\ (min)} & \textbf{Accuracy} \\
        \midrule
        FT full &  90   & 13012  & 40.34  & 0.90    \\ 
        LoRA    &  80   & 11823 & 32.10 & 0.90  \\ 
        \ours   &  70   & 5341 & 29.30 & 0.90 \\
        LoRA + \ours  & \textbf{50}  & \textbf{4243}  & 21.20 & \textbf{0.89} \\
        \bottomrule
    \end{tabular}
    }
\end{table}

\begin{table}[t]
    \caption{Combining LoRA + \ours for fine-tuning  \vit on \food dataset}
    \label{table:LoRA_combine_food}
    \centering
    \scalebox{0.8}
    {
    \begin{tabular}{l | c c c c}
    \hline
    \toprule
        \textbf{Method} & \parbox{1.5cm}{\centering \textbf{Compute} \\ (PFLOPs)} & \parbox{1.5cm}{\centering \textbf{Memory} \\ (MB)} & \parbox{1.5cm}{\centering \textbf{Time} \\ (min)} & \textbf{Accuracy} \\
        \midrule
        FT full & 150 & 13012 & 64.37 & \textbf{0.90}    \\ 
        LoRA  & 109  & 11823 & 48.35 & 0.89  \\ 
        \ours &  107 & 5451 & 48.42 & 0.90 \\
        LoRA + \ours  & \textbf{95}   & \textbf{4243}  & 41.00 & \textbf{0.90} \\
        \bottomrule
    \end{tabular}
    }
\end{table}

\section{Compute Budget Allocation} \label{sec:appb}
\begin{figure*}[t]
    \centering
    \begin{tabular}{ccc}
        \begin{subfigure}{0.3\textwidth}
            \centering
            \includegraphics[width=\textwidth]{images/distributions/deit_t_food.pdf}
            \caption{ \deitt on \food}
        \end{subfigure} &
        \begin{subfigure}{0.3\textwidth}
            \centering
            \includegraphics[width=\textwidth]{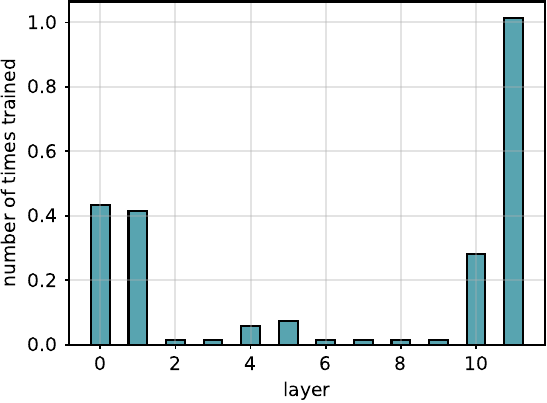}
            \caption{ \deitt on \cifar}
        \end{subfigure} &
        \begin{subfigure}{0.3\textwidth}
            \centering
            \includegraphics[width=\textwidth]{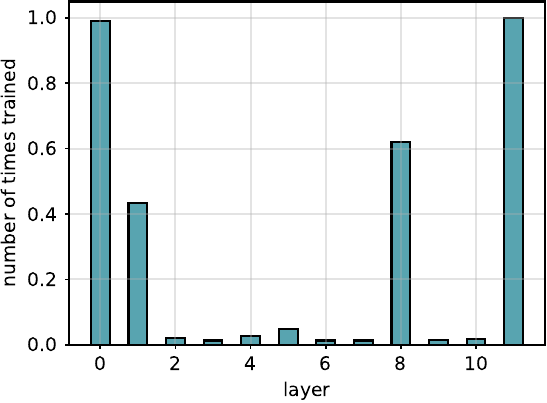}
            \caption{ \deitt on \flowers}
        \end{subfigure} \\
        
        \begin{subfigure}{0.3\textwidth}
            \centering
            \includegraphics[width=\textwidth]{images/distributions/deit_s_food.pdf}
            \caption{ \deits on \food}
        \end{subfigure} &
        \begin{subfigure}{0.3\textwidth}
            \centering
            \includegraphics[width=\textwidth]{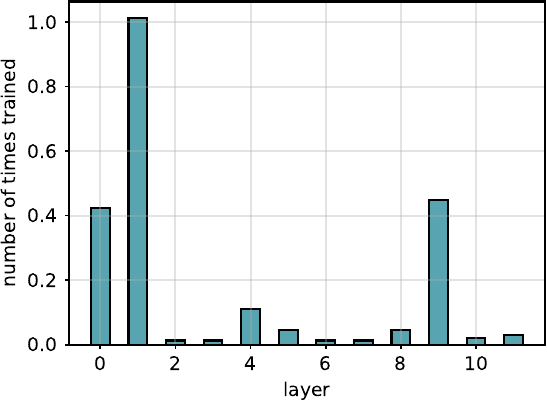}
            \caption{ \deits on \cifar}
        \end{subfigure} &
        \begin{subfigure}{0.3\textwidth}
            \centering
            \includegraphics[width=\textwidth]{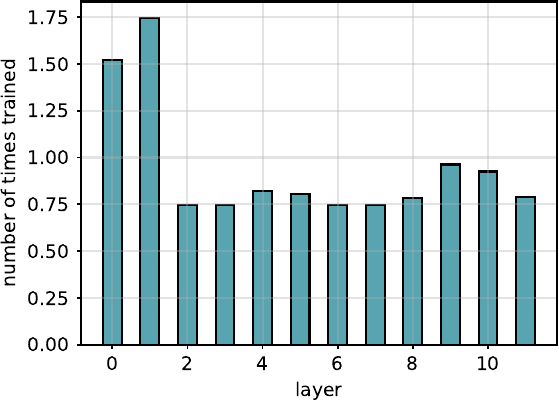}
            \caption{ \deits on \flowers}
        \end{subfigure} \\
        
        \begin{subfigure}{0.3\textwidth}
            \centering
            \includegraphics[width=\textwidth]{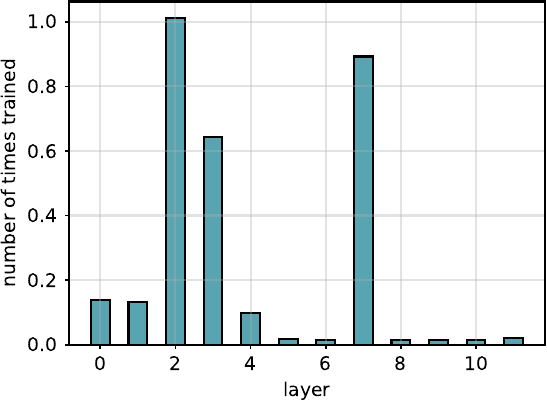}
            \caption{ \vit on \food}
        \end{subfigure} &
        \begin{subfigure}{0.3\textwidth}
            \centering
            \includegraphics[width=\textwidth]{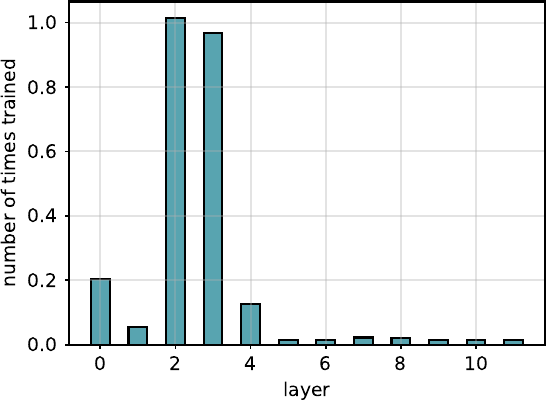}
            \caption{ \vit on \cifar}
        \end{subfigure} &
        \begin{subfigure}{0.3\textwidth}
            \centering
            \includegraphics[width=\textwidth]{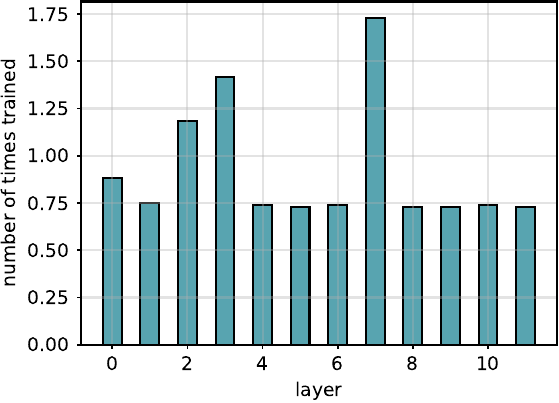}
            \caption{ \vit on Flowes102}
        \end{subfigure}
    \end{tabular}
    \caption{Final assignment on budgets for training on different datasets and models.}
    \label{fig:all_budgets_final}
\end{figure*}

\begin{figure*}[ht]
    \centering
    \begin{tabular}{ccc}
        \begin{subfigure}{0.3\textwidth}
            \centering
            \includegraphics[width=\textwidth]{images/distribution_dynamic/budgets_deit_t_food.pdf}
            \caption{ \deitt on \food}
        \end{subfigure} &
        \begin{subfigure}{0.3\textwidth}
            \centering
            \includegraphics[width=\textwidth]{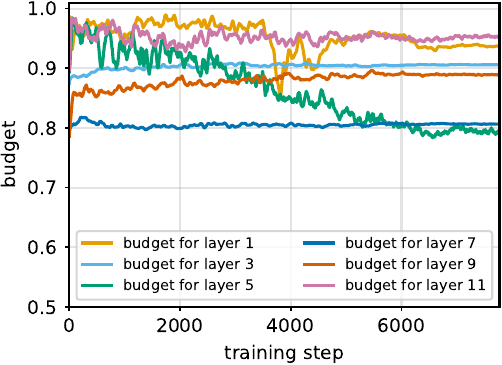}
            \caption{ \deitt on \cifar}
        \end{subfigure} &
        \begin{subfigure}{0.3\textwidth}
            \centering
            \includegraphics[width=\textwidth]{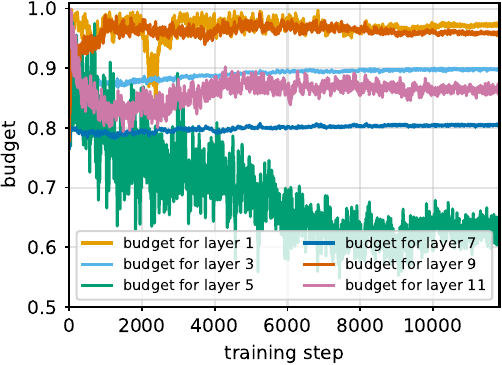}
            \caption{ \deitt on \flowers}
        \end{subfigure} \\
        
        \begin{subfigure}{0.3\textwidth}
            \centering
            \includegraphics[width=\textwidth]{images/distribution_dynamic/budgets_deit_s_food.pdf}
            \caption{ \deits on \food}
        \end{subfigure} &
        \begin{subfigure}{0.3\textwidth}
            \centering
            \includegraphics[width=\textwidth]{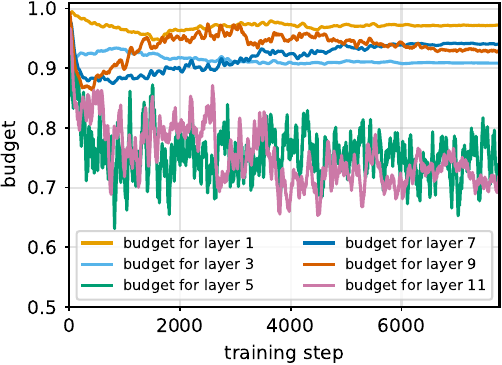}
            \caption{ \deits on \cifar}
        \end{subfigure} &
        \begin{subfigure}{0.3\textwidth}
            \centering
            \includegraphics[width=\textwidth]{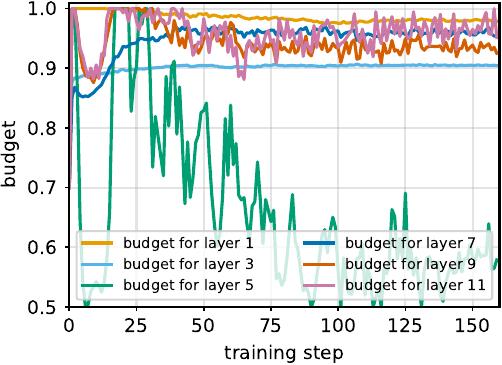}
            \caption{ \deits on \flowers}
        \end{subfigure} \\
        
        \begin{subfigure}{0.3\textwidth}
            \centering
            \includegraphics[width=\textwidth]{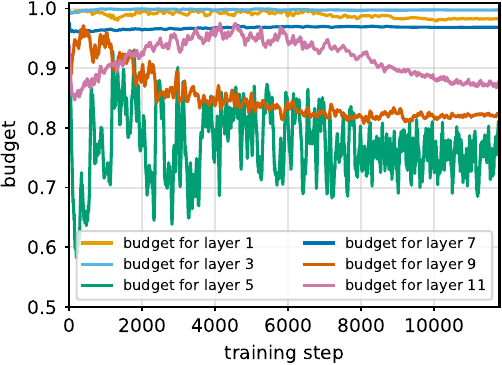}
            \caption{ \vit on \food}
        \end{subfigure} &
        \begin{subfigure}{0.3\textwidth}
            \centering
            \includegraphics[width=\textwidth]{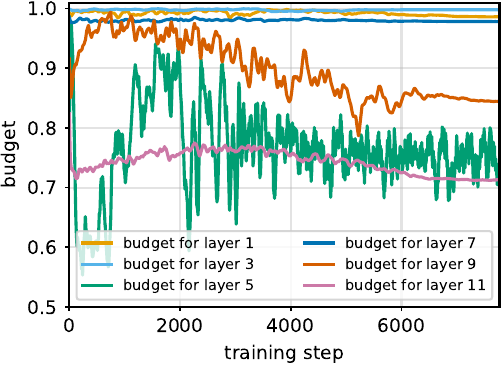}
            \caption{ \vit on \cifar}
        \end{subfigure} &
        \begin{subfigure}{0.3\textwidth}
            \centering
            \includegraphics[width=\textwidth]{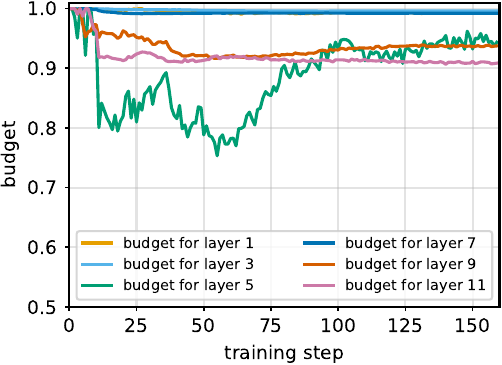}
            \caption{ \vit on \flowers}
        \end{subfigure}
    \end{tabular}
    \caption{Compute budget assignment during fine-tuning.}
    \label{fig:all_budgets_ft}
\end{figure*}

In \Cref{fig:all_budgets_final} we show the total number of times each layer is chosen during fine-tuning. We see that for  \deits and  \vit the compute budget is more evenly distributed across layers, and initial layers are usually picked more frequently. In the smaller  \deitt, on the other hand, the budget is allocated mainly to initial and last layers, and the distribution results more peaked. In \Cref{fig:all_budgets_ft} we show the budget assignment during fine-tuning. We observe that central layers usually exhibit a higher variance of assigned compute budget. In  \deits and  \vit we observe a pattern where deeper layers are assigned high budget in the first iterations, and the budget is then decreased along fine-tuning. For the initial layers on the other hand, the budget is usually constantly high across the entire fine-tuning. 

\section{Implementation details} \label{sec:appe}
In the following, we provide details on the implementation and fine-tuning hyper-parameters. We fine-tune the three models using Adam optimizer \cite{KingBa15} with a batch size of $128$. After an initial grid search, we set the learning rate to $0.0001$ for all methods but LoRA, where we found $0.001$ to be more effective. We use two random augmentations picked between horizontal or vertical flipping and random cropping for all datasets and runs. All images are resized to $224 \times 224$, that is the required input size for the pretrained models.
We download the weights of the models (pre-trained on ImageNet \cite{ILSVRC15}) from Timm Image models \cite{rw2019timm}. 
During fine-tuning we keep track of FLOPs, peak memory and training time on an NVidia RTX4090 GPU. To measure FLOPs and memory load, we use PyTorch MACs counter and memory allocation tools \cite{paszke2017automatic}. Finally, we provide the Python code used for experiments.

\begin{table*}[t]
    \caption{Comparison of Fine-tuning \deitt \deits and  \vit on \flowers dataset.}
    \label{table:flowers102}
    \centering
    \scalebox{0.8}
    {
    \begin{tabular}{l | c c c c | c c c c | c c c c}
    \hline
    \toprule
    \textbf{Model} & \multicolumn{4}{c|}{\textbf{Deit-t-16}} & \multicolumn{4}{c|}{\textbf{Deit-s-16}} & \multicolumn{4}{c}{\textbf{ \vit}}  \\[0.2cm]
    \textbf{} & \parbox{1.2cm}{\centering \textbf{Compute} \\  (PFLOPs)} & \parbox{1.2cm}{\centering \textbf{Memory} \\  (MB)} & \parbox{1.2cm}{\centering \textbf{Time} \\  (min)} & \textbf{Accuracy} & \parbox{1.2cm}{\centering \textbf{Compute} \\  (PFLOPs)} & \parbox{1.2cm}{\centering \textbf{Memory} \\ (MB)} & \parbox{1.2cm}{\centering \textbf{Time} \\  (min)} & \textbf{Accuracy} & \parbox{1.2cm}{\centering \textbf{Compute} \\  (PFLOPs)} & \parbox{1.2cm}{\centering \textbf{Memory} \\ (MB)} & \parbox{1.2cm}{\centering \textbf{Time} \\  (min)} & \textbf{Accuracy} \\
        \midrule
        FT head  & 0.12 & 2389  & \textbf{1.11}& 0.51 & \textbf{0.3} & 4751  & 1.5 & 0.54 & \textbf{1.0} & 9657 & \textbf{3.58} &  0.73\\ 
        FT full  & 0.14 & 3267  & 2.12& 0.72 & 0.5 & 7186  & 2.4 & \textbf{0.89} & 1.5 & 13012& 7.18 &  \textbf{0.97}\\ 
        FT top-3 & 0.12 & 2525  & 1.19& 0.55 & 0.4 & 5202 & 1.2 & 0.77  & 1.2 & 7096 & 4.25 &  0.90\\ 
        ToME \citenump{bolya2022tokenmerging}    & 0.13 & 2983  & 1.39& \textbf{0.88} & 0.5 & 2134  & 3.0  & 0.88 & 1.3 & \textbf{4493} & 5.31 &  0.95\\ 
        BSR  \citenump{bsr}    & \textbf{0.11} & \textbf{2000}  & 1.15& 0.73 & 0.39& \textbf{3520} & \textbf{1.1} & 0.76  & \textbf{1.0} & 6900 & 4.12 &  0.90\\ 
        LoRA  \citenump{hu2022lora}   & \textbf{0.11} & 2400  & 2.19& 0.56 & 0.5 & 7256 & 2.5  & 0.71 & \textbf{1.0} & 11823 & 7.44 & \textbf{0.97}\\ 
        \ours (ours)& \textbf{0.8} & \textbf{3820}  & \textbf{1.39} & \textbf{0.79} & \textbf{0.3} & \textbf{3523}  & \textbf{2.2} & \textbf{0.9} & \textbf{1.5} & \textbf{5341} & \textbf{5.31} & \textbf{0.97} \\ 
    \bottomrule
    \end{tabular}
    }
\end{table*}

\end{document}